\documentclass[11pt]{article}

\usepackage[preprint]{acl}

\usepackage{times}
\usepackage{latexsym}

\usepackage[T1]{fontenc}

\usepackage[utf8]{inputenc}

\usepackage{booktabs}

\usepackage{microtype}

\usepackage{inconsolata}

\usepackage{graphicx}

\usepackage{hyperref}
\usepackage{url}
\usepackage{graphicx}
\usepackage{float}
\usepackage{enumitem}
\usepackage{xspace}
\usepackage{amsmath}
\usepackage{amssymb}
\usepackage{graphicx}
\usepackage[dvipsnames]{xcolor}
\usepackage{cleveref}
\usepackage{listings}
\usepackage{multirow}
\usepackage{rotating}
\usepackage{subcaption}
\usepackage{tabularx}
\usepackage{booktabs}
\usepackage{enumitem}
\usepackage{fontawesome5}

\newcommand{\llada}{LLaDA-8B\xspace}
\newcommand{\mmada}{MMaDA-8B\xspace}
\newcommand{\dream}{Dream-7B\xspace}
\newcommand{\llama}{LLama-3.1-8B\xspace}

\newcommand{\numstd}[2]{$#1 \scriptscriptstyle \pm #2$}

\newcommand\blfootnote[1]{%
  \begingroup
  \renewcommand\thefootnote{}\footnote{#1}%
  \addtocounter{footnote}{-1}%
  \endgroup
}

\definecolor{darkblue}{rgb}{0, 0, 0.5}
\hypersetup{colorlinks=true, citecolor=darkblue, linkcolor=darkblue, urlcolor=darkblue}

\title{Attention Sinks in Diffusion Language Models}

\author{
  \textbf{Maximo Eduardo Rulli\textsuperscript{\dag}\textsuperscript{\(\ast\)}}
  \textbf{Simone Petruzzi\textsuperscript{\dag}\textsuperscript{\(\ast\)}}
  \textbf{Edoardo Michielon\textsuperscript{\ddag}}
\\
  \textbf{Fabrizio Silvestri\textsuperscript{\dag}}
  \textbf{Simone Scardapane\textsuperscript{\dag}}
  \textbf{Alessio Devoto\textsuperscript{\dag}}
\\
\\
  \textsuperscript{\dag}Sapienza University of Rome
  \textsuperscript{\ddag} Fastweb
\\
  \\
\footnotesize
{\faGithub}\hspace{0.5em}\texttt{\url{https://github.com/Maximo-Rulli/dlms-sinks}}
}

\begin{document}
\maketitle
\begin{abstract}
Masked Diffusion Language Models (DLMs) have recently emerged as a promising alternative to traditional Autoregressive Models (ARMs).
DLMs employ transformer encoders with bidirectional attention, enabling parallel token generation while maintaining competitive performance.
Although their efficiency and effectiveness have been extensively studied, the internal mechanisms that govern DLMs remain largely unexplored.
In this work, we conduct an empirical analysis of DLM attention patterns, focusing on the attention sinking phenomenon, an effect previously observed in various transformer-based architectures.
Our findings reveal that DLMs also exhibit attention sinks, but with distinct characteristics.
First, unlike in ARMs, the sink positions in DLMs tend to shift throughout the generation process, displaying a dynamic behaviour.
Second, while ARMs are highly sensitive to the removal of attention sinks, DLMs remain robust: masking sinks leads to only a minor degradation in performance.
These results provide new insights into the inner workings of diffusion-based language models and highlight fundamental differences in how they allocate and utilize attention compared to autoregressive models.
\end{abstract}

\section{Introduction}
\blfootnote{\textsuperscript{\(\ast\)} Equal contribution.\\ Correspondence to: rulli.2154435@studenti.uniroma1.it, alessio.devoto@uniroma1.it}
Large Language Models (LLMs) have driven a paradigm shift across numerous scientific and industrial domains, demonstrating remarkable capabilities in language understanding, generation, and reasoning~\citep{gpt4, claude, qwen3, llama4}. This rapid progress is rooted in the transformer architecture and the attention mechanism~\citep{transformer}. While attention is a critical aspect of the transformer's effectiveness, it also gives rise to complex and often non-intuitive emergent phenomena.

One of the most striking traits of these behaviours is the "attention sink"~\citep{streamingllm, offbyone}.
This consists in the fact that, in most autoregressive models (ARMs), a small subset of tokens consistently receives a disproportionate amount of attention from other tokens in the sequence.
The pattern is not limited to language, and similar patterns have been observed in Vision Transformers~\citep{registers} and encoder-only transformers~\citep{rusciogeometric}, suggesting it may be a fundamental property of attention-based deep networks.
\begin{figure}[t]
    \centering
    \includegraphics[width=0.48\textwidth]{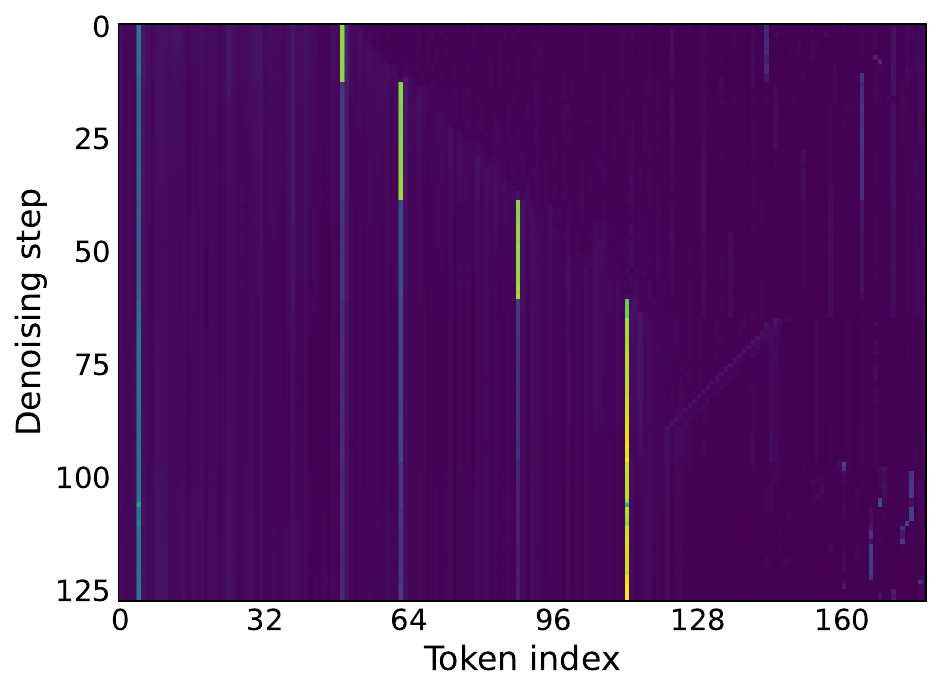}    
    \caption{Incoming attention scores for each token in \llada~\citep{llada} across denoising steps. Unlike autoregressive models, DLMs exhibit attention sinks that shift across the sequence as tokens are progressively unmasked.}
    \label{fig:llada_sink_first_page}
\end{figure}
Recently, masked Diffusion Language Models (DLMs) have emerged as an alternative to the dominant autoregressive paradigm~\citep{llada, dream, geminidiff, inceptiondiffusion, mmada, wangrevolutionizing, songseed, sahoosimple, lladamoe, longllada}. 
Unlike Autoregressive Models (ARMs), which generate text strictly from left to right, DLMs iteratively refine a fully masked sequence through successive denoising steps~\citep{llada, dream, mmada}. 
Generation is based on the unmasking of an initial fully masked sequence of tokens, that the model progressively "denoises" over multiple steps to produce a coherent fully unmasked output. Crucially, DLMs employ a bidirectional attention mechanism. 
%
%
While this bidirectional information flow is key to their parallel, non-causal generation process, the precise impact of this architecture on the inner workings of DLMs remains largely unexplored.

In this work, we present an empirical study of attention patterns in DLMs, focusing specifically on the attention sink phenomenon.
We analyse three state-of-the-art open-source masked DLMs: \dream~\citep{dream}, a model initialized from a pre-trained ARM; \llada~\citep{llada}, a large-scale model trained from scratch; and \mmada~\citep{mmada}, a multimodal DLM trained from \llada. 
Our analysis reveals that DLMs do exhibit attention sinks, but these sinks possess unique dynamic properties rarely seen in their autoregressive counterparts. 
Unlike the static attention sinks well-documented in ARMs, most of the sinks in DLMs are unstable and their position actively shifts across the iterative denoising process. 
Additionally, while ARMs are extremely sensitive to removing the sink tokens, we find that DLMs are significantly more robust to this intervention. 
We attribute this property to their decoding strategy that unmasks only the tokens with highest probabilities in the sequence, and the lack of a causal mask that limits the attention interaction among tokens. 
%
%
To summarize, our primary contributions are the following:
\begin{itemize}[leftmargin=*]
\item We conduct an empirical study on attention patterns in DLMs, and provide empirical evidence that attention sinks consistently emerge in these models.
\item We characterize the dynamic properties of these sinks, showing they can disappear and shift positions during inference, and we introduce a metric to track their intensity and location across denoising steps.
\item We investigate how model performance is affected by removing sinks, and show DLMs are robust to sink masking.
\end{itemize}
\begin{figure*}[t]
    \centering    
    \begin{subfigure}{0.28\textwidth}
        \centering
        \includegraphics[width=\textwidth]{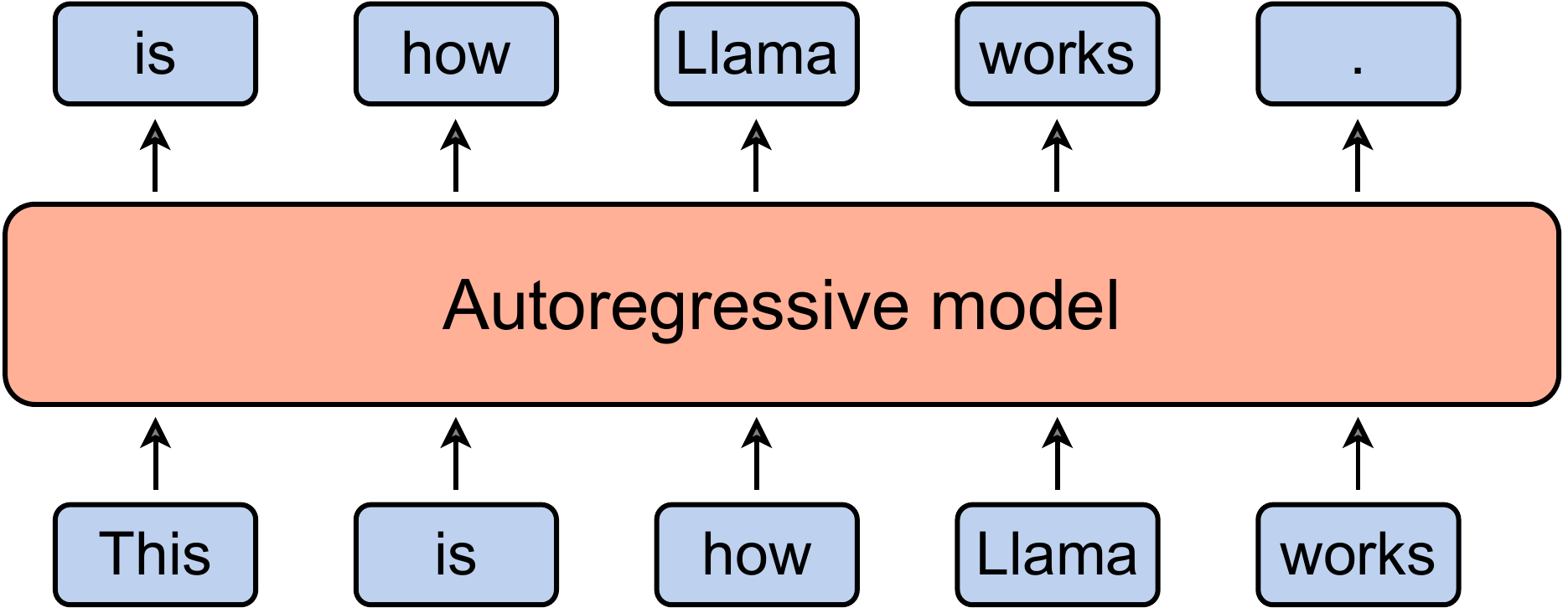}
        \caption{In ARMs, each token is used to predict the next one in the sequence. Then, the predicted token is appended to the input sequence and fed again to the model, in a left to right fashion.}
        \label{fig:sub:ref}
    \end{subfigure}
    \hfill 
    \begin{subfigure}{0.33\textwidth}
        \centering
        \includegraphics[width=\textwidth]{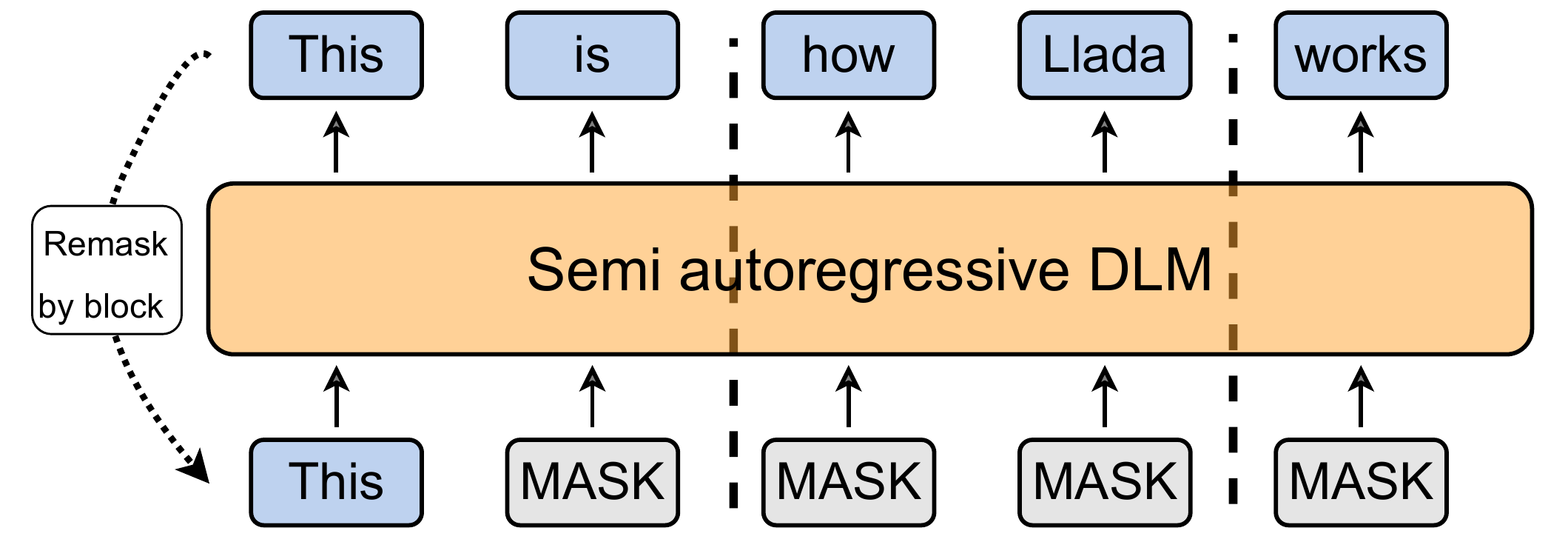}
        \caption{In \llada and \mmada a sequence of \texttt{[MASK]} tokens is passed as input to the model. The model then performs N denoising steps, gradually unmasking tokens inside each block, before proceding to the next block from left to right.}
        \label{fig:sub:mmada}
    \end{subfigure}
    \hfill 
    \begin{subfigure}{0.33\textwidth}
        \centering
        \includegraphics[width=\textwidth]{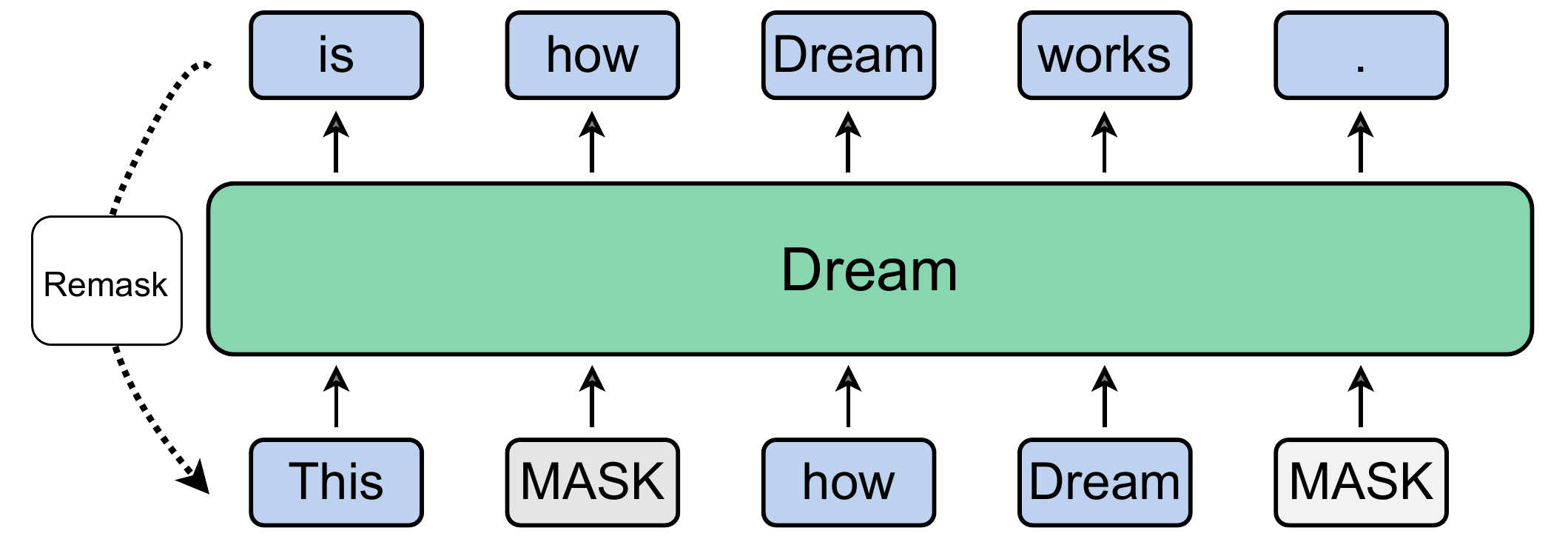}
        \caption{\dream is initialized from an ARM. Each token is used to predict the following one. At each inference step the entire input masked sequence is passed to the model. Unlike in other DLMs, \dream uses the current token to predict the next one.}
        \label{fig:sub:dream}
    \end{subfigure}
    \caption{Snapshot of an inference step for different language models. ARMs and \dream predict the next token, while \mmada and \llada predict the current one. \mmada and \llada perform semi-autoregressive block decoding, where only tokens in the current block are unmasked, while \dream may unmask a token at any position.}
    \label{fig:inference_comparison}
\end{figure*}
\section{Related Work}
\subsection{Diffusion Language Models}
%
%
%
Language modelling has traditionally been dominated by autoregressive models that generate text sequentially, one token at a time. 
While this paradigm has proven highly successful, DLMs have emerged as an alternative, offering token generation through an iterative denoising processes with potential efficiency advantages~\citep{dlmssurvey,wufast, kimany, longllada, libeyond, wufast-2}. 
Some applications of diffusion to language modelling operate in continuous space, first embedding discrete tokens into continuous vectors, applying diffusion-based denoising, and then mapping back to discrete tokens~\citep{li2022diffusion, strudel2022self, gongdiffuseq, dieleman2022continuous}. 
While theoretically elegant, this approach introduces additional complexity in handling the discrete nature of language. A more direct approach emerged with discrete diffusion models, which operate directly on token vocabularies~\citep{austin2021structured, gongscaling, hoogeboom2021argmax, campbell2022continuous}. 
Starting from fully masked sequences of \texttt{[MASK]} tokens, these models iteratively predict and refine tokens through a process reminiscent of BERT-style masked language modelling~\citep{he2023diffusionbert, gong2025scalingdiffusionlanguagemodels}. 
Several works~\citep{austin2021structured, he2023diffusionbert, gong2025scalingdiffusionlanguagemodels} have adopted this paradigm but faced significant scaling challenges, remaining limited in size while autoregressive models scaled to billions of parameters.  
Recently, discrete DLMs have gained traction thanks to open-source models like \dream~\citep{dream},  \mmada~\citep{mmada} and \llada~\citep{llada, lladamoe, longllada}, which have successfully scaled to 7 billion parameters and beyond, narrowing the performance gap with ARMs.

In this work, we investigate attention patterns in large discrete DLMs, that operate directly on the vocabulary space.
\subsection{Attention Sink in Transformers}
Attention Sink refers to the common phenomenon observed in transformers where a small subset of tokens consistently receives a disproportionate amount of attention from other tokens in the sequence. This behaviour was initially discovered in \citet{streamingllm}, and leveraged for efficiency. 
After this, other works have then explored the sink phenomenon, characterizing properties of sink tokens like high $L_2$ norm in the hidden state activations~\citep{massive, cancedda} or low $L_2$ norm in the key projection~\citep{knorm, guattention}.
Similar properties have been also observed in the vision domain~\citep{registers}. 
Several works have attempted to explain the emergence of attention sinks in transformers. \citet{when} offers an empirical study of how attention sinks manifest in transformer models, specifically focusing on ARMs. \citet{glasses, barbero2025llmsattendtoken} and \citet{pappone2025attentionsinks} investigate the phenomenon analytically and show how attention sinks act as a bias for ARMs and can mitigate information over-squashing. 
Finally, ~\citet{rusciogeometric} analyses attention sinks from a geometric perspective, and shows that they emerge to establish stable coordinate systems in the model's high-dimensional latent space.
While these works analyse sinks in both decoder and encoder transformers, we are the first to observe and investigate this phenomenon in the context of DLMs.
\section{Background on Masked Discrete Diffusion}
\label{sec:background}
Traditional ARMs model the probability of a text sequence $\mathbf{x} = (x_1, x_2, \dots, x_L)$ of length $L$ by decomposing the joint probability into a product of conditional probabilities, generated in a strict, left-to-right order~\citep{jelinek1980interpolated,bengio2000neural}.
This decomposition is given by:
\begin{equation}
    p(\mathbf{x}) = p(x_1) \prod_{i=2}^{L} p(x_i | x_1, \dots, x_{i-1})  
\end{equation}
where $x_i$ is the token at position $i$, and $p(x_i | x_1, \dots, x_{i-1})$ is the probability of the current token conditioned only on all preceding tokens. Masked discrete DLMs offer a non-autoregressive, parallel alternative. 
Instead of generating tokens one by one, they model a Markov diffusion process over discrete token sequences.
This consists of two complementary phases: a fixed \textbf{forward corruption process} and a learned \textbf{reverse denoising process}. 
The forward process systematically corrupts a clean data sequence $\mathbf{x}_0$ (the original text) over a series of time steps $t \in \left[ 0, T\right]$ by progressively replacing tokens with a special mask token \texttt{[MASK]}.
Starting with the clean sequence $\mathbf{x}_0$, a noisy sequence $\mathbf{x}_t$ at time step $t$ is generated by a Markov transition $q(\mathbf{x}_t | \mathbf{x}_{t-1})$. 
The marginal distribution of a token $\mathbf{x}_t^i$ at time $t$ conditioned on its clean version $\mathbf{x}_0^i$ is defined by a masking schedule $\alpha_t \in [0, 1]$. 
The complete forward process is the joint distribution over all intermediate noisy states, a product of the Markov transitions:
\begin{equation}
    q(\mathbf{x}_{1:T} | \mathbf{x}_0) = \prod_{t=1}^{T} q(\mathbf{x}_t | \mathbf{x}_{t-1})
\end{equation}
In the denoising process, a model $p_{\boldsymbol{\theta}}$, parametrized by $\theta$, reverses this noising process, generating new data from a fully masked sequence $\mathbf{x}_T$ back to a clean sequence $\mathbf{x}_0$. 
More specifically, reverse transition $p_{\boldsymbol{\theta}}(\mathbf{x}_{t-1} | \mathbf{x}_t)$ is parameterized by the model, which is trained to estimate the true reverse conditional probability $q(\mathbf{x}_{t-1} | \mathbf{x}_t)$. 
In practice, the model $p_{\boldsymbol{\theta}}$ is often trained to predict the clean data $\mathbf{x}_0$ from the noisy input $\mathbf{x}_t$ at a given time $t$, and this prediction is then used to approximate the reverse transition. 
The model output is a distribution over the original tokens, from which the next, less-noisy state $\mathbf{x}_{t-1}$ is sampled.

In this work we consider three masked discrete DLMs: \llada~\citep{llada}, \mmada~\citep{mmada} and \dream~\citep{dream}. 
\llada and \mmada are trained from scratch, with a masked language modelling loss where a token $x_i$ is masked during the forward process, and the model learns to predict the token itself ($x_i \rightarrow $ \texttt{[MASK]} $\rightarrow x_i$). 
At inference time, \llada and \mmada use semi-autoregressive block diffusion, where the input sequence is divided into blocks, and the model gradually unmasks  all tokens inside the corresponding block in a left-to-right manner~\citep{arriolablock}, (see \Cref{fig:inference_comparison}). 
\dream, on the other hand, is initialized from an autoregressive model to leverage the pretrained weights and its training objective employs a "shift operation"~\citep{dream, gong2025scalingdiffusionlanguagemodels}. 
More specifically, when a token $x_i$ is masked, \dream is trained to predict $x_{i+1}$, similarly to an autoregressive model ($x_i \rightarrow $ \texttt{[MASK]} $\rightarrow x_{i+1}$). In \Cref{fig:inference_comparison} we provide a comparison and visual explanation of how the different types of inference are implemented.
\section{Analysis of Attention Sinks in Masked Diffusion Language Models}
\label{sec:analysis}
\begin{figure}[t]
    \centering
    \includegraphics[width=0.9\linewidth]{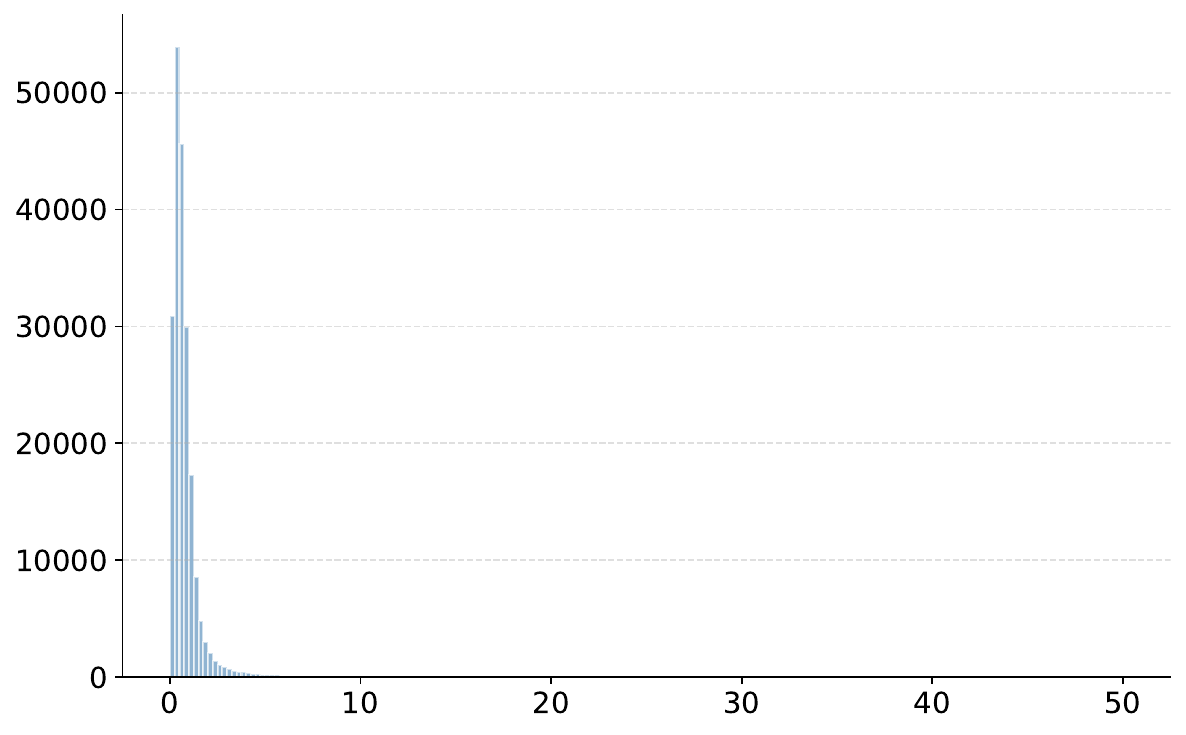}
    \caption{Distribution of attention scores in \llada~\citep{llada} across denoising steps. Only a few tokens, the attention sinks, receive a very high attention score, while the majority of tokens in the sequence have scores close to zero.}
    \label{fig:lladahist}
\end{figure}

Previous work has shown that attention sinks emerge in most transformer-based architectures, regardless of the data domain and training strategy~\citep{when, rusciogeometric, streamingllm, registers}. 
Attention sinks are characterized by the disproportionate attention score they receive from all the tokens in the sequence, and can be easily identified as vertical bright lines in attention maps (like the one we show in \Cref{fig:llada_sink_first_page}). 
To validate the presence of attention sinks in DLMs, we first analyse the distribution of attention scores in \llada and show it in \Cref{fig:lladahist}. We see that only a few tokens, the sinks, capture a very high attention score consistently. Similar patterns emerge for \dream and \mmada (see \Cref{app:hists}). 
We now define a metric to characterize and locate attention sinks in DLMs. 
\begin{figure*}[t] 
    \centering

    \begin{subfigure}{0.48\textwidth}
        \centering
        \includegraphics[width=0.49\textwidth]{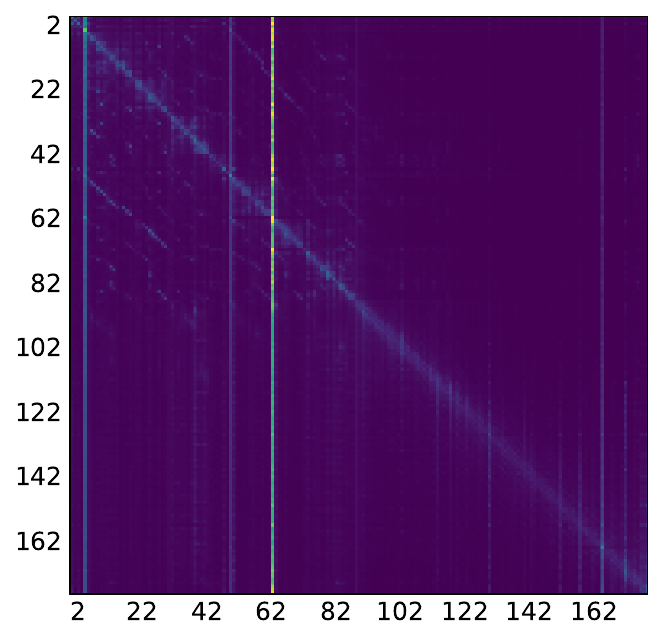}
        \hfill 
        \includegraphics[width=0.49\textwidth]{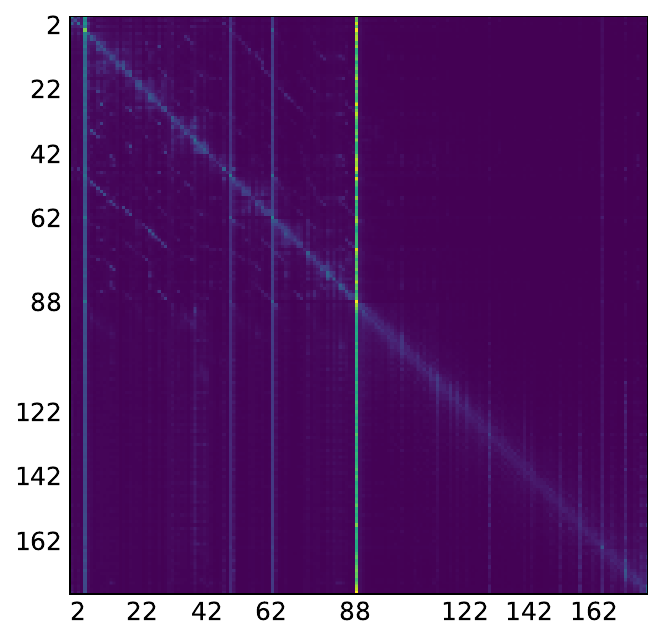}
        \caption{\textbf{Moving sink in \llada}. Attention plots at step 38 (Left) and step 39 (Right). The sink shifts from position 62 to 88 after one denoising step.}
        \label{fig:llada_moving_sink}
    \end{subfigure}%
    \hfill 
    \begin{subfigure}{0.48\textwidth}
        \centering
        \includegraphics[width=0.49\textwidth]{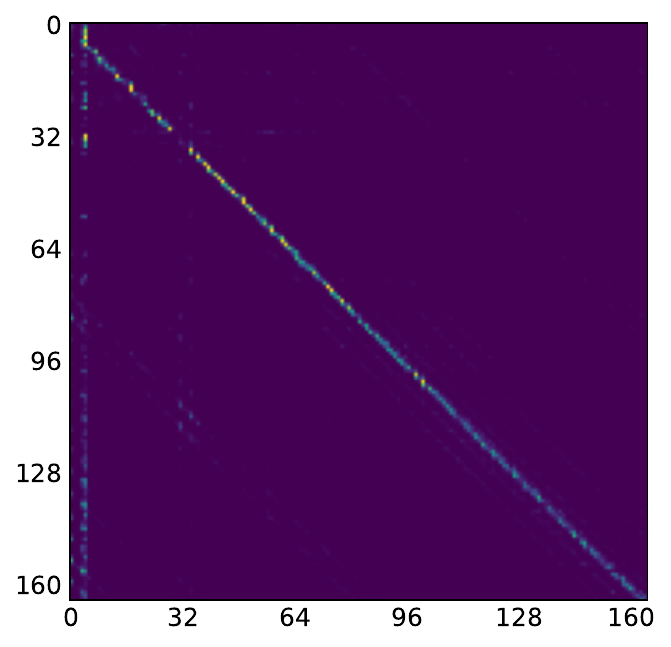}
        \hfill
        \includegraphics[width=0.49\textwidth]{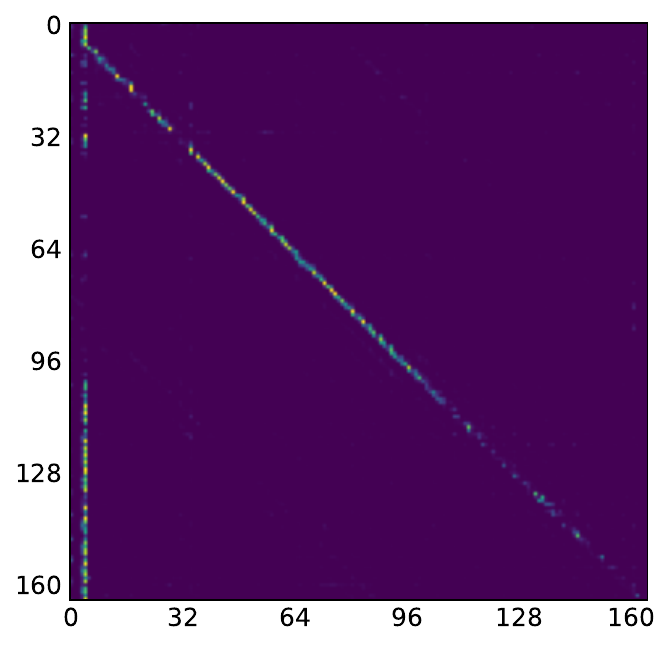}
        \caption{\textbf{Moving sink in \mmada}. Attention at step 36 (Left) and step 37 (Right). Observe that this sink absorbs the self-attention from each of the tokens paying it attention.}
        \label{fig:mmada_moving_sink}
    \end{subfigure}

    \caption{Moving sink in \llada, and \mmada.}
    \label{fig:combined_sinks}
\end{figure*}
\subsection{Definition of Attention Sink} Consider an encoder-only transformer model. For a single attention head $h$ and layer $l$, we have that the attention score is defined as:
\begin{equation*}
    A_{ij} = \text{softmax}_j\left(\frac{q_i^\top k_j}{\sqrt{d}}\right)
\end{equation*}
where $q_i \text{ and } k_j$ are the query and key projections for token $i \text{ an } j$ respectively, and $A_{ij}$ represents the amount of attention that token $i$ pays to token $j$. 
In a DLM attention is bidirectional, and we obtain a distribution of attention scores across the entire sequence \textit{at each denoising step}. 
Given the attention scores, we define the cumulative attention score for a token $j$ as the average attention it receives from all tokens in a specific denoising step $t$:
\begin{equation*}
    \bar{A}^{(t,l,h)}_{j} = \frac{1}{S} 
    \sum_{i=1}^{S}
    A_{ij}^{(t,l,h)}
\end{equation*}
where $S$ is the sequence length, and $A_{i,j}^{(t,l,h)}$ represents the attention score from token $i$ to token $j$ at denoising step $t$, in head $h$ of layer $l$. 
We then identify attention sinks as tokens that receive a cumulative attention score substantially larger than the average.
%
%

\textbf{Attention Sink.} We formally define a token $j$ at a specific denoising step $t$, in head $h$ of layer $l$ to be a \textbf{sink token}, if its cumulative attention score exceeds the average cumulative attention score of all other tokens by at least a threshold $\epsilon$:
\begin{equation}
\label{eq:metric}
    j \text{ is a sink token if } \bar{A}^{(t,l,h)}_{j} > \frac{1}{S-1} \sum_{k \neq j} \bar{A}^{(t,l,h)}_{k} + \epsilon
\end{equation}
This definition ensures that sink tokens represent significant outliers in the attention distribution. 
In all our experiments we use $\epsilon=3$, which we selected to filter out at least the 96\% of tokens in sequence, and empirically showed a sufficient robustness to detect sinks while also serving as a filter for tokens that did not exhibit a sink characteristic. We further discuss the value of $\epsilon$ in \Cref{app:epsilon}.
%
%
%
%
%
\subsection{Sink Patterns}
\label{sec:patterns}
Our analysis reveals that DLMs exhibit distinct types of attention sinks with unique dynamic properties not observed in ARMs. 
We find that sinks do not necessarily appear in the beginning of the sentence, but also show up in the middle or towards the end, which is possible as attention in DLMs is bidirectional. 
Along with the typical static sink that is frequently observed in ARMs, we identify a new kind of attention sinks that we call \textbf{moving sinks}.

\begin{figure}[t]
    \centering
    \includegraphics[width=0.8\linewidth]{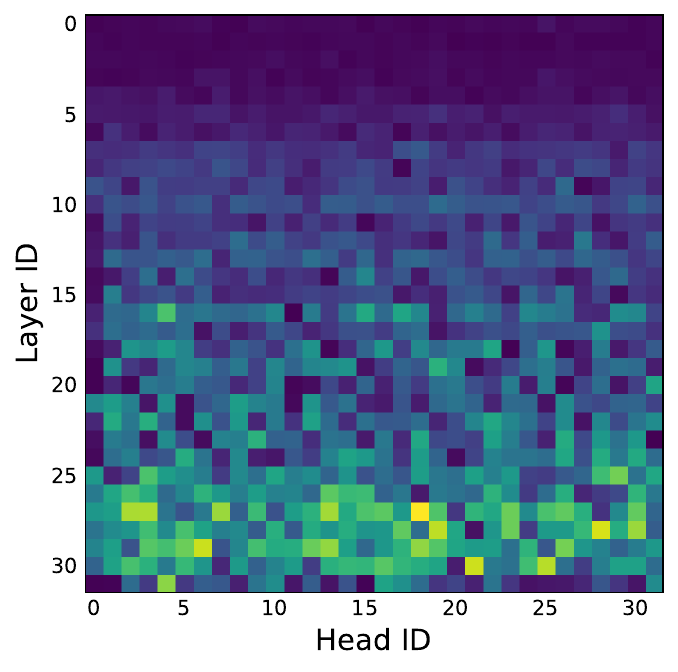}
    \caption{\textbf{Cumulative attention score for \llada's sink across heads and layers}. The variation of the model's main sink token is displayed across the different heads and layers, averaged through time. Brighter colours indicate higher attention score. In later layers there are usually fewer sinks and the attention score is therefore higher, as it is shared among fewer sink tokens.}
    \label{fig:llada_deep}
\end{figure}

\begin{figure*}[t] 
    \centering
    \begin{subfigure}{0.23\textwidth}
        \centering
        \includegraphics[width=\textwidth]{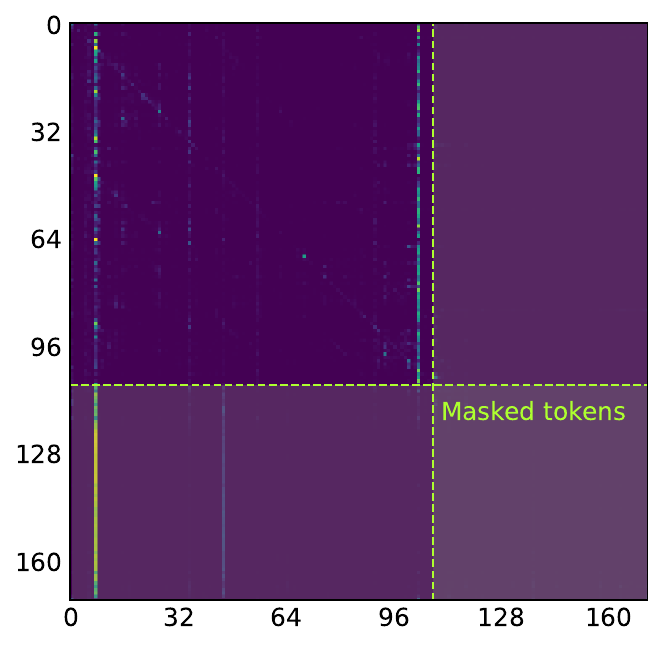}
        \caption{} 
        \label{fig:complementary_sink}
    \end{subfigure}%
    \hfill 
    \begin{subfigure}{0.72\textwidth}
        \centering
        \includegraphics[width=0.32\textwidth]{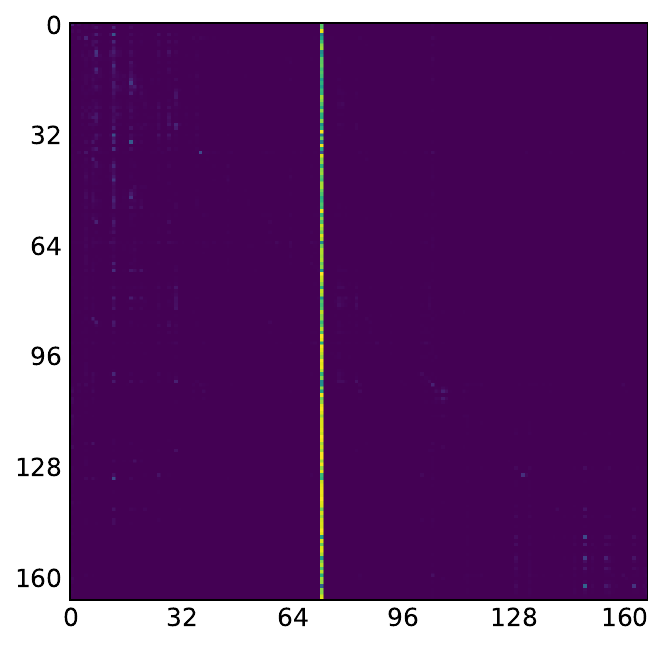}
        \hfill
        \includegraphics[width=0.32\textwidth]{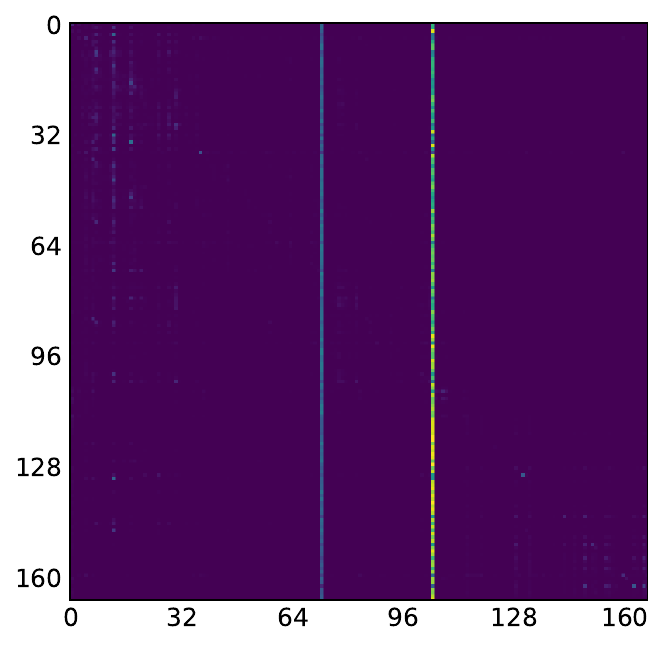}
        \hfill
        \includegraphics[width=0.32\textwidth]{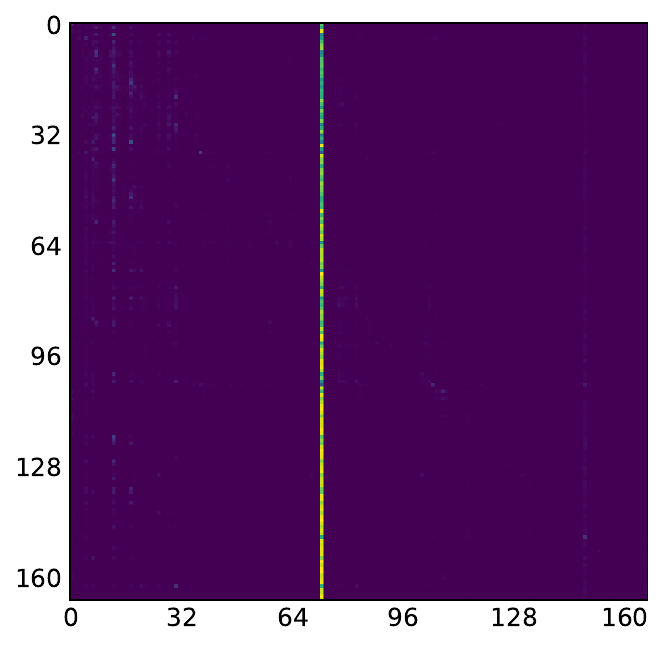}
        \caption{} 
        \label{fig:erratic_sink}
    \end{subfigure}
    \caption{\textbf{Different types of moving sinks in \llada.} \textbf{(a)} A particular kind of moving sink in which attention is split according to token type. Some heads exhibit this behaviour in which the masked tokens heavily attend to a specific sink, while the unmasked ones are more concentrated on another one. This heatmap is from step 32, at the precise end of a block, explaining why we have a perfect line separating all the unmasked and masked tokens. \textbf{(b)} A sink appears at step 96 but suddenly disappears at step 97.}
    \label{fig:compl_erratic}
\end{figure*}

\begin{figure*}[t] 
    \centering
    \begin{subfigure}{0.48\textwidth}
        \centering
        \includegraphics[width=0.49\textwidth]{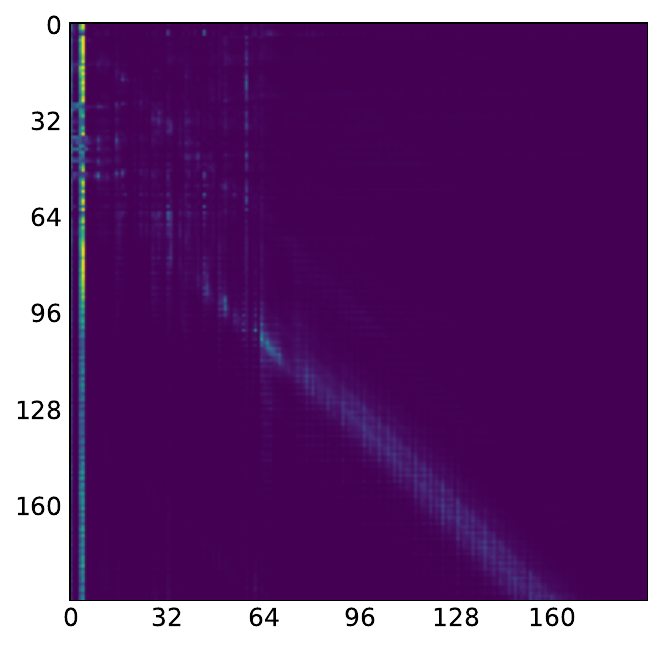}
        \hfill 
        \includegraphics[width=0.49\textwidth]{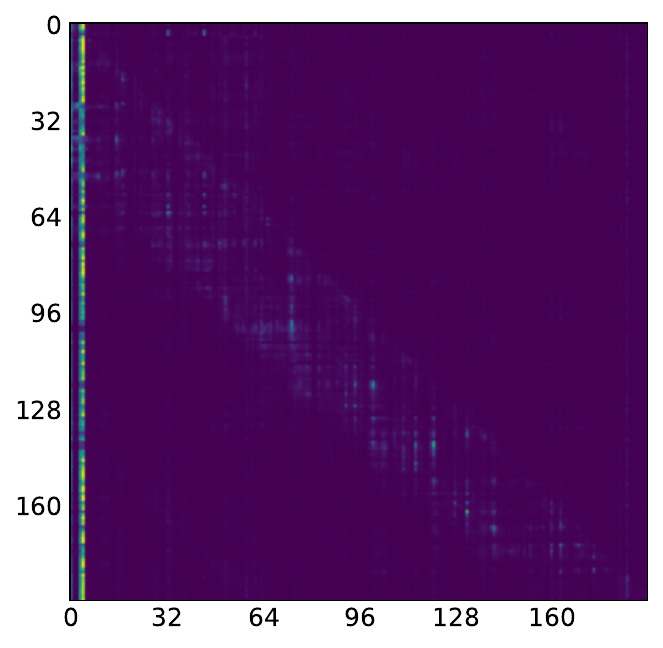}
        \caption{\textbf{Fixed sink in \mmada}. \mmada often exhibits a static sink at the beginning of the sequence. In different denoising steps (0 and 127), the sink stays consistently at the beginning of the sequence.}
        \label{fig:mmada_fixed}
    \end{subfigure}%
    \hfill 
    \begin{subfigure}{0.48\textwidth}
        \centering
        \includegraphics[width=0.49\textwidth]{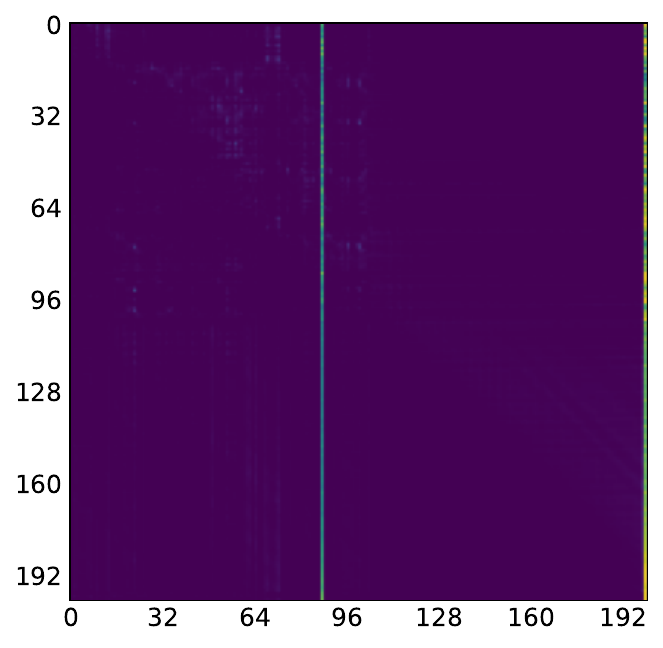}
        \hfill
        \includegraphics[width=0.49\textwidth]{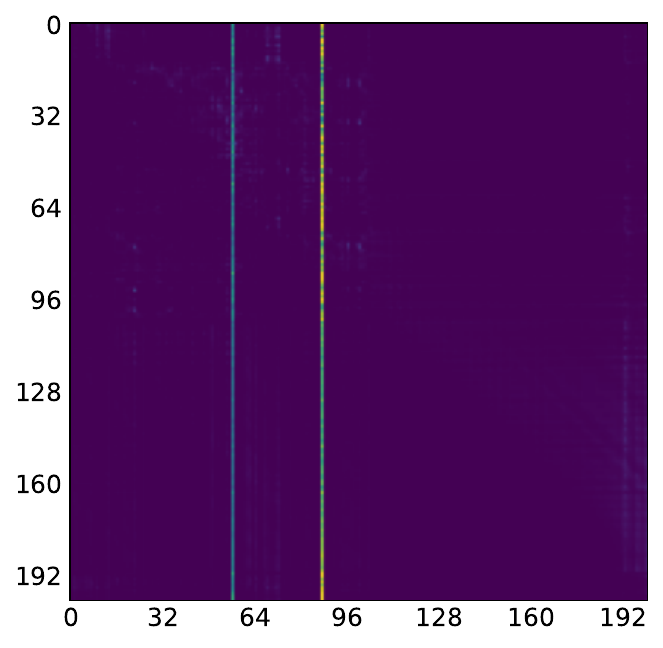}
        \caption{\textbf{Moving sinks in \dream} typically shift from right to left. The sink moving is on step 32 (Left) and at the rightmost position. While at step 33 (Right) the sink has moved towards the centre.} 
        \label{fig:dream_moving}
    \end{subfigure}
    \caption{Fixed sink in \mmada and moving sink in \dream.}
    \label{fig:dream_mmada}
\end{figure*}

\textbf{Moving sinks} appear at different positions during denoising and exhibit widely different patterns according to layer depth and backbone model. 
Moving sinks are not consistent across diffusion steps, i.e. they do not remain at the same position across all diffusion steps and may move or even vanish throughout the denoising process. We show an example in \Cref{fig:llada_moving_sink}.
%
%
We now analyse how attention sinks appear in the considered pre-trained models.
\paragraph{\llada} exhibits diverse moving sink patterns with consistency across different sequences. 
Moving sinks often remain at a specific position for some consecutive denoising steps, before vanishing. 
Nonetheless, we also find some edge cases in which the moving sinks behave extremely unstably, as we see in \Cref{fig:erratic_sink}, where a sink appears for only one timestep before vanishing on the next one.
As we progress to deeper layers, the number of sinks decreases, converging to one or two sinks per layer, as we show in \Cref{fig:llada_deep}. 
The deepest layers showcase a particular type of moving sinks, where masked and unmasked tokens maintain separate attention sinks, and switch gradually. We show an example of this phenomenon in \Cref{fig:complementary_sink}. 
Notably, \llada demonstrates a strong semantic basis for sink selection as sinks consistently form on punctuation marks (periods, commas), whitespace, and end-of-sequence tokens. 
This pattern suggests that \llada, trained from scratch as a diffusion model, developed semantically-aware attention mechanisms that identify structurally important tokens as reference points for attention.
\paragraph{\dream} showcases a sink behaviour that follows primarily a positional rather than a semantic pattern. 
Unlike \llada, \dream's sinks often originate at the rightmost masked token and shift leftward as tokens are progressively unmasked, regardless of the token content, as we show in \Cref{fig:dream_moving}.
This right-to-left migration is most prominent in early layers and creates a dynamic attention flow that follows the unmasking frontier. This positional nature of \dream's sinks likely stems from its initialization from a pre-trained autoregressive model. 
The inherited representations may be less refined for bidirectional attention, causing the model to rely on positional cues rather than semantic content for sink formation. 
%
\dream's positional bias represents a difference from \llada's semantic approach and suggests that initialization strategy and positional embeddings significantly influences attention organization in diffusion models~\citep{rusciogeometric}.
%
%
%
%
%
%
\paragraph{\mmada} presents the most stable sink behaviour among the three models, with sinks that are generally static and less frequent. 
When sinks do manifest they often remain fixed at their initial positions throughout the entire generation process, as we show in \Cref{fig:mmada_fixed}. 
The model exhibits minimal moving sinks, with most layers showing no clear sink patterns at all. 
This stability contrasts with the dynamic patterns in \llada and \dream, potentially reflecting \mmada's different multimodal training data.
The static nature of \mmada's sinks more closely resembles traditional autoregressive models, though the bidirectional attention mechanism still allows for unique patterns not possible in causal models. 
For instance, in \Cref{fig:mmada_moving_sink} we show that  a considerable amount of tokens shift their attention towards an already unmasked token from one step to the other.
\begin{figure}[t]
    \centering
    \includegraphics[width=0.9\linewidth]{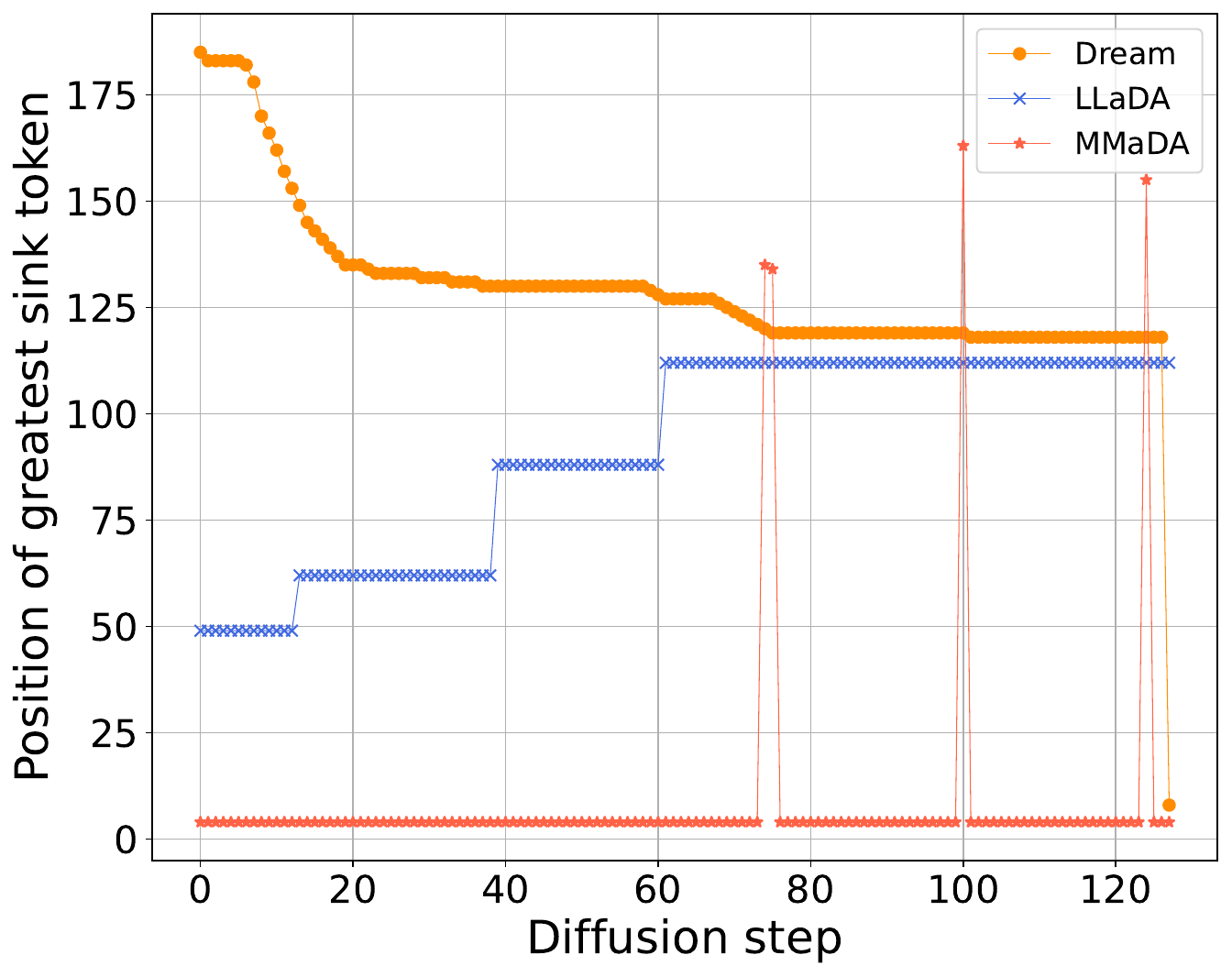}
    \caption{\textbf{Example of how sinks move over time.} The largest sink from each model's specific heads is selected at each iteration. See how the attention shifts according to the explained phenomena. Note that these are sinks for a specific head of the model and not the actual averaged one.}
    \label{fig:overview}
\end{figure}

In \Cref{fig:overview} we show an example of the phenomena described previously for the different models. We select a specific head from each model and compare the position of the largest sink detected by our metric. 
We observe that while \mmada exhibits a mostly static sinking behaviour, sinks tend to shift position in \dream and \llada. More specifically, we observe that in \llada the sink tends to shift right as more blocks are denoised, while it moves from right to left in \dream.
\begin{table*}[t]
  \centering
  \renewcommand{\arraystretch}{1.2}
  \setlength{\tabcolsep}{7pt}
  \small
  \resizebox{\textwidth}{!}{%
  \begin{tabular}{llcccc}
    \toprule
    \textbf{Dataset} & \textbf{Sinks} 
    & \textsc{Dream-7B} [\citenum{dream}]
    & \textsc{LLaDA-8B} [\citenum{llada}]
    & \textsc{MMaDA-8B} [\citenum{mmada}]
    & \textsc{Llama-3.1-8B} [\citenum{llama3}] \\
    \midrule
    \multirow{4}{*}{\textbf{GSM8K}} 
      & Unmasked             & \numstd{0.82}{0.01} & \numstd{0.76}{0.01} & \numstd{0.54}{0.01} & \numstd{0.85}{0.01} \\
      & Masked $ \epsilon_0 $  & \numstd{0.79}{0.01} & \numstd{0.75}{0.01} & \numstd{0.53}{0.01} &  \numstd{0.02}{0.00} \\
      & Masked $ \epsilon_1 $  & \numstd{0.78}{0.01} & \numstd{0.73}{0.01} & \numstd{0.54}{0.01} & \numstd{0.02}{0.00} \\
      & Masked $ \epsilon_2 $ & \numstd{0.75}{0.01} & \numstd{0.55}{0.01} & \numstd{0.37}{0.01} & \numstd{0.01}{0.03} \\
    \midrule
    \multirow{4}{*}{\textbf{HumanEval}} 
      & Unmasked             & \numstd{0.60}{0.03} & \numstd{0.37}{0.03} & \numstd{0.16}{0.02} & \numstd{0.66}{0.04} \\
      & Masked $ \epsilon_0 $  & \numstd{0.64}{0.03} & \numstd{0.37}{0.03} & \numstd{0.16}{0.03} & \numstd{0.00}{0.00}\\
      & Masked $ \epsilon_1 $  & \numstd{0.61}{0.03} & \numstd{0.39}{0.03} & \numstd{0.18}{0.03} & \numstd{0.00}{0.00} \\
      & Masked $ \epsilon_2 $ & \numstd{0.57}{0.03} & \numstd{0.35}{0.03} & \numstd{0.09}{0.02} & \numstd{0.00}{0.00} \\
    \bottomrule
  \end{tabular}}
  \caption{\textbf{Performance of DLMs and ARMs under attention sink masking.} Thresholds $\epsilon_0$, $\epsilon_1$, and $\epsilon_2$ correspond to masking 1, 5, and 10 top-ranked attention sinks, respectively. While \llama exhibits severe degradation when masking a single sink token, \llada, \dream, and \mmada maintain competitive performance across masking settings, suggesting that parallel inference in DLMs provides robustness to attention sink removal.}
  \label{tab:model_comparison}
\end{table*}

\subsection{Robustness of DLMs to Masking Sinks}
\label{sec:exp}
Previous studies have demonstrated that attention sinks play a crucial role in transformer-based models, with their removal typically causing catastrophic performance degradation~\citep{streamingllm, when, glasses}. 
However, given that attention sinks in DLMs exhibit markedly different and more dynamic patterns compared to ARMs, we investigate whether DLMs demonstrate similar sensitivity to sink masking during generation.

We evaluate the three DLM variants — \llada, \dream, and \mmada — on both coding and mathematical reasoning tasks using the GSM8K~\citep{gsm8k} and HumanEval~\cite{humaneval} datasets. 
GSM8K contains grade-school level math word problems, while HumanEval comprises programming problems designed to evaluate code generation and reasoning capabilities.
We evaluate each model in two configurations: (1) the original, unmodified model, and (2) with the top-K sink tokens masked. To determine which tokens to mask, we analyse the attention map from step $t-1$ across each head and layer. We identify the top-K sink tokens based on our metric (Equation \ref{eq:metric}) and mask them by setting their attention to $-\infty$ during the computation of step $t$. We vary the threshold parameter $\epsilon$, where smaller values result in masking a larger proportion of sinks. Specifically, we select $\epsilon_0$, $\epsilon_1$ and $\epsilon_2$ to mask the top $1, 5 \text{ and } 10$ sinks respectively.

Surprisingly, the tested DLMs exhibit only modest performance degradation when sinks are masked (Table~\ref{tab:model_comparison}). For all the tested DLMs, masking one sink leads to a degradation in performance smaller than 1\%. Substantial degradation occurs only when $\epsilon$ is decreased further to mask $10$ sinks, and mostly in \mmada.
To make a fair comparison with ARMs, we conduct a similar experiment with \llama, where instead of analyzing the entire attention map to determine the token to mask at the next iteration, we focus on the attention distribution of the last token. We adopt this approach because ARMs, due to their causal nature, cannot 'shift' their attention as flexibly as DLMs. In contrast to DLMs, applying this masking procedure to \llama results in severe performance drops even when masking a single sink token, confirming prior findings that ARMs are highly sensitive to attention sink removal~\citep{streamingllm, when}.
We hypothesize that this increased robustness stems from the parallel inference mechanism inherent to DLMs, which may provide alternative attention pathways when primary sinks are unavailable. We explore this hypothesis further in Section~\ref{sec:robust}.
\paragraph{Implementation details.} We evaluate our models in PyTorch~\citep{torch} using the checkpoints released on Hugging Face transformers~\citep{transformers} and the official \verb|lm evaluation harness| scripts~\citep{eval-harness}. 
We use the same hyper-parameters specified in the respective original papers. For \llada, we use a block size of 32 and a generation length of 256 tokens for GSM8K and 512 for HumanEval. 
For \dream, which does not use semi-autoregressive block generation, we adjust only the generation length and diffusion step parameters according to the original settings. 
We successfully reproduce the reported results for \llada, \dream, and \llama using these configurations. However, we were unable to reproduce the original results for \mmada despite following the published implementation details, and we therefore report our own evaluation results for this model.
Throughout our analysis, we employ $\epsilon=3$ for sink detection, a threshold that empirically balances robust sink identification with the exclusion of non-sink tokens.
\section{Discussion}
\subsection{Dynamic Sinks and Positional Encoding}
Recent work on encoder-only models notes that attention sinks can shift usually around special markers like \texttt{[CLS]} or \texttt{[EOS]} and connects this behaviour to the use of absolute positional embeddings~\citep{rusciogeometric}. 
However, we find that DLMs, despite using Rotary Positional Embeddings~(RoPE, ~\citealt{rope}), show extremely varied and dynamic sink patterns, including sinks that move and others that split attention between masked and unmasked tokens. These appear all over the text sequence, often on important structural tokens (like punctuation). We show relative frequencies of sink tokens in Table \ref{tab:token_frequencies}.
The emergence of sink tokens on semantic and formatting markers suggests that the sinking behaviour is driven not only by the positional encoding or token index in the sequence~\citep{rusciogeometric, barbero2025llmsattendtoken}, but also by training dynamics and frequency of the token in the training corpus~\cite{massive, magikarp}.
\subsection{Robustness to Masking Sinks}
\label{sec:robust}
A notable result from \Cref{sec:exp} is that DLMs keep working, although with a drop in performance, even when we mask their attention sinks, which would cause an ARM to fail completely. 
We believe this robustness comes from the bidirectional attention and the iterative denoising process working together to create stability that ARMs lack. 
In ARMs, attention is causal, and the sink token is usually a single, static anchor, that all future tokens rely on. 
The next token to predict is therefore usually highly dependent on the sink, and cutting its attention score causes the model to fail.

However, the bidirectional attention in DLMs lets every token see the full context at every denoising step. 
Additionally, at each step all tokens are considered for unmasking, and only the ones with highest probability (i.e., where the model is most confident) are actually unmasked. 
This iterative denoising process might ensure higher stability: when a sink is masked, the model likely becomes less confident about those tokens that are highly affected by the sink, and therefore not consider them for unmasking.

\begin{table}[htbp]
\centering

\setlength{\tabcolsep}{5pt} 
\scalebox{0.95}{
\begin{tabular}{llr}
\toprule
\textbf{Model} & \textbf{Token} & \textbf{Freq.} \\
\midrule
\multirow{5}{*}{LLaDA} 
  & \texttt{Ċ} & 0.368 \\
  & \texttt{<|mdm\_mask|>} & 0.366 \\
  & \texttt{.} & 0.050 \\
  & \texttt{<|start\_header\_id|>} & 0.008 \\
  & \texttt{(white-space)} & 0.008 \\
\midrule
\multirow{5}{*}{Dream} 
  & \texttt{<|mask|>} & 0.321 \\
  & \texttt{Ċ} & 0.090 \\
  & \texttt{.} & 0.046 \\
  & \texttt{,} & 0.044 \\
  & \texttt{<|endoftext|>} & 0.040 \\
\midrule
\multirow{5}{*}{MMaDA} 
  & \texttt{Ċ} & 0.180 \\
  & \texttt{<|mdm\_mask|>} & 0.166 \\
  & \texttt{<|end\_header\_id|>} & 0.160 \\
  & \texttt{<|startoftext|>} & 0.108 \\
  & \texttt{istant} & 0.066 \\
\bottomrule
\end{tabular}
}
\caption{\textbf{Relative Frequency of Top-5 Sink Tokens}. Relative frequency of the top-5 most frequent tokens detected as sinks by our metric. Across all models the \texttt{Ċ} (a token representing white-spaces) and \texttt{[MASK]} token are the ones at which the models most often sink their attention. We note that in MMaDA the \texttt{istant} token comes from \texttt{assistant} , which is part of the prompt template. Frequencies computed across 30 generations with generation length of 128, 128 denoising steps and block size of 32.}
\label{tab:token_frequencies}
\end{table}

\subsection{Long Context Modelling}
In ARMs, attention sinks have been proven to act as a tool to control over-mixing and avoid representation collapse, especially in long contexts~\citep{barbero2025llmsattendtoken, oversquashing}.
However, attention sinks in ARMs are usually present only at the beginning of the sequence and represent a single point of reference for the entire generation. 
In contrast, DLMs offer a flexible inference and their sinks often shift position during generation. 
%
By dynamically directing attention to tokens that are currently most important for the ongoing prediction, DLMs might be able to maintain strong, long-range connections more effectively than ARMs that rely on a single, fixed bottleneck for information. 
Having the ability to access sinks at the end of the sequence might represent and advantage for long reasoning and planning tasks~\citep{ye2024diffusion,ye2024beyond, ye2025implicit}, where the model needs a reference anchor in the future instead of the usual static one at the beginning of the sequence.
Additionally, for very long context generation in real-world deployment scenarios, sinks represent a single point of weakness. When the context exceeds the available GPU memory, the oldest part, typically including the \texttt{[BOS]} token, must be discarded. However, discarding sinks in ARMs has been shown to be catastrophic for downstream performance. 
DLMs on the other hand mitigate this limitation. Their moving sinks, which often appear in the future relative to the current generation step, allow the model to discard the past context without significant performance degradation.
\section{Conclusion}
\label{sec:conclusion}
We presented the first empirical analysis of attention sinks in Diffusion Language Models, showing that they consistently emerge but behave differently from those in autoregressive models. 
In DLMs, sinks are dynamic, often shifting across denoising steps and aligning with semantic or structural tokens rather than fixed positions.
Moreover, DLMs remain remarkably robust to sink masking, suggesting that their bidirectional and iterative generation distributes attention more evenly and avoids reliance on single anchor tokens. 
These findings reveal that diffusion models organize attention through flexible mechanisms, offering new insights into their internal dynamics and interpretability.
\section{Future Work}
While our empirical analysis offers a general overview of sink behaviour in DLMs, it also raises several open questions.
First, it remains unclear what type of information the model stores in the sinks that correspond to future positions. A promising direction to investigate this would be a mechanistic analysis, for instance using the Logit Lens~\citep{logitlens}.

Second, it is worth exploring whether sinks could be exploited for acceleration or compression, similar to their original use case in~\citep{streamingllm}.

Finally, although we observed several sink behaviours (e.g., \Cref{fig:complementary_sink}), we did not attempt to provide a detailed explanation of these phenomena. While such an investigation would be valuable, it would require an interpretability-focused study, which lies beyond the scope of this primarily empirical work.
\section{Limitations}
\label{sec:limitations}
While we conducted an extensive study across three DLMs, our analysis is limited to instruct models, as we did not perform experiments on their corresponding base versions. Furthermore, we focused on attention sinks in pre-trained models and did not explore how modifications to the training procedure might influence their behaviour, an aspect that has recently been investigated for ARMs by \citet{offbyone, gptoss}.
\section{Acknowledgements}
We thank Fastweb S.p.a. for providing the computational resources used in this paper. We also thank Jary Pomponi, Pasquale Minervini and Emile van Krieken for helpful discussions and valuable feedback.
\bibliography{references}
\appendix
\begin{figure*}[t]
    \begin{subfigure}{0.4\textwidth}
        \centering
        \includegraphics[width=\textwidth]{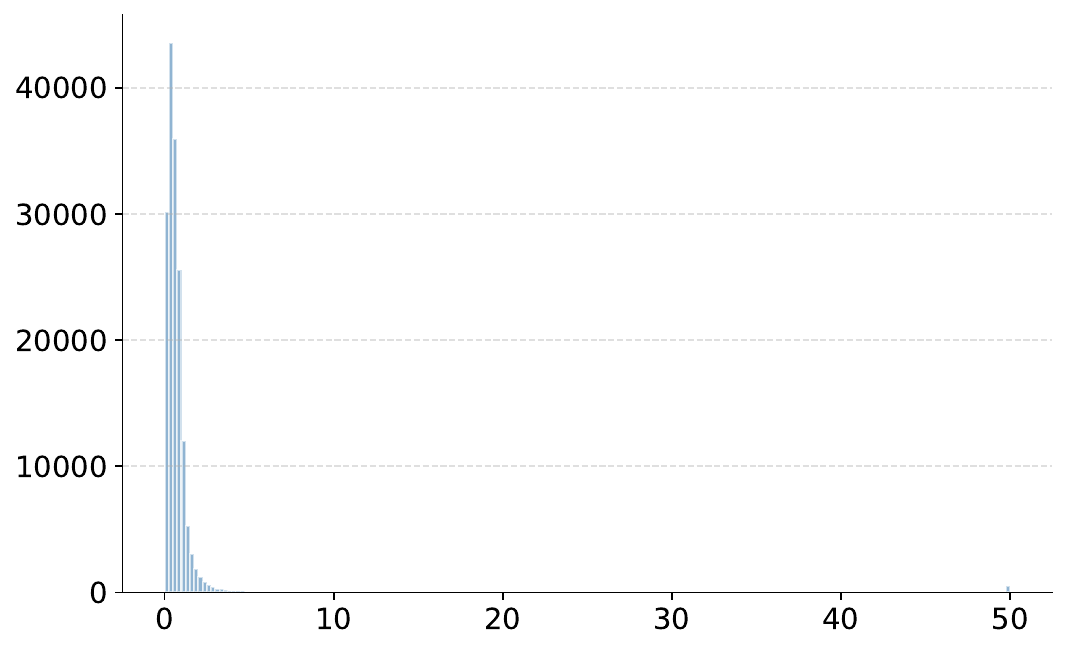}
    \end{subfigure}
    \hfill 
    \begin{subfigure}{0.4\textwidth}
        \centering
        \includegraphics[width=\textwidth]{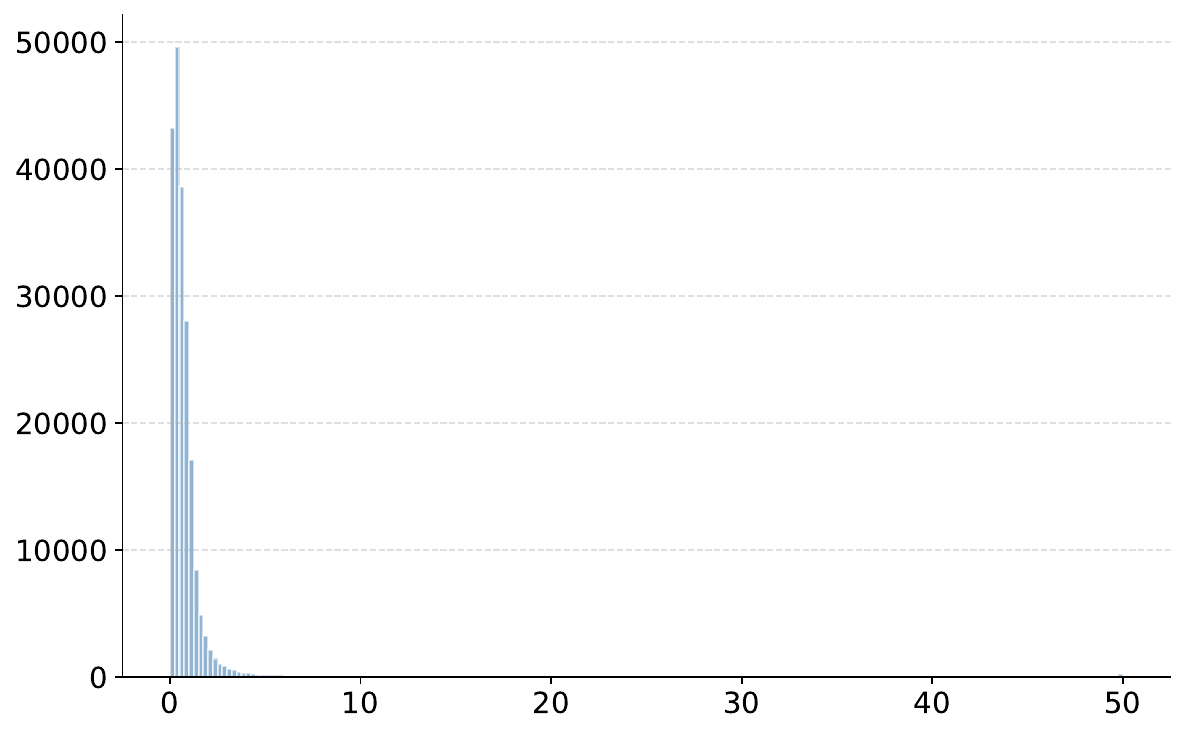}
    \end{subfigure}
    \caption{Distribution of attention scores in \dream and \mmada}
    \label{fig:hist_incoming_llada}
\end{figure*}
\section{Additional plots}
\label{app:hists}
In \Cref{fig:hist_incoming_llada} we show additional plots of attention score distribution, displaying how a only a few tokens, the sinks, receive a disproportionate high attention score.
\section{Selection of Sink Threshold}
\label{app:epsilon}
\begin{figure}[H]
    \centering
        \centering
        \includegraphics[width=\linewidth]{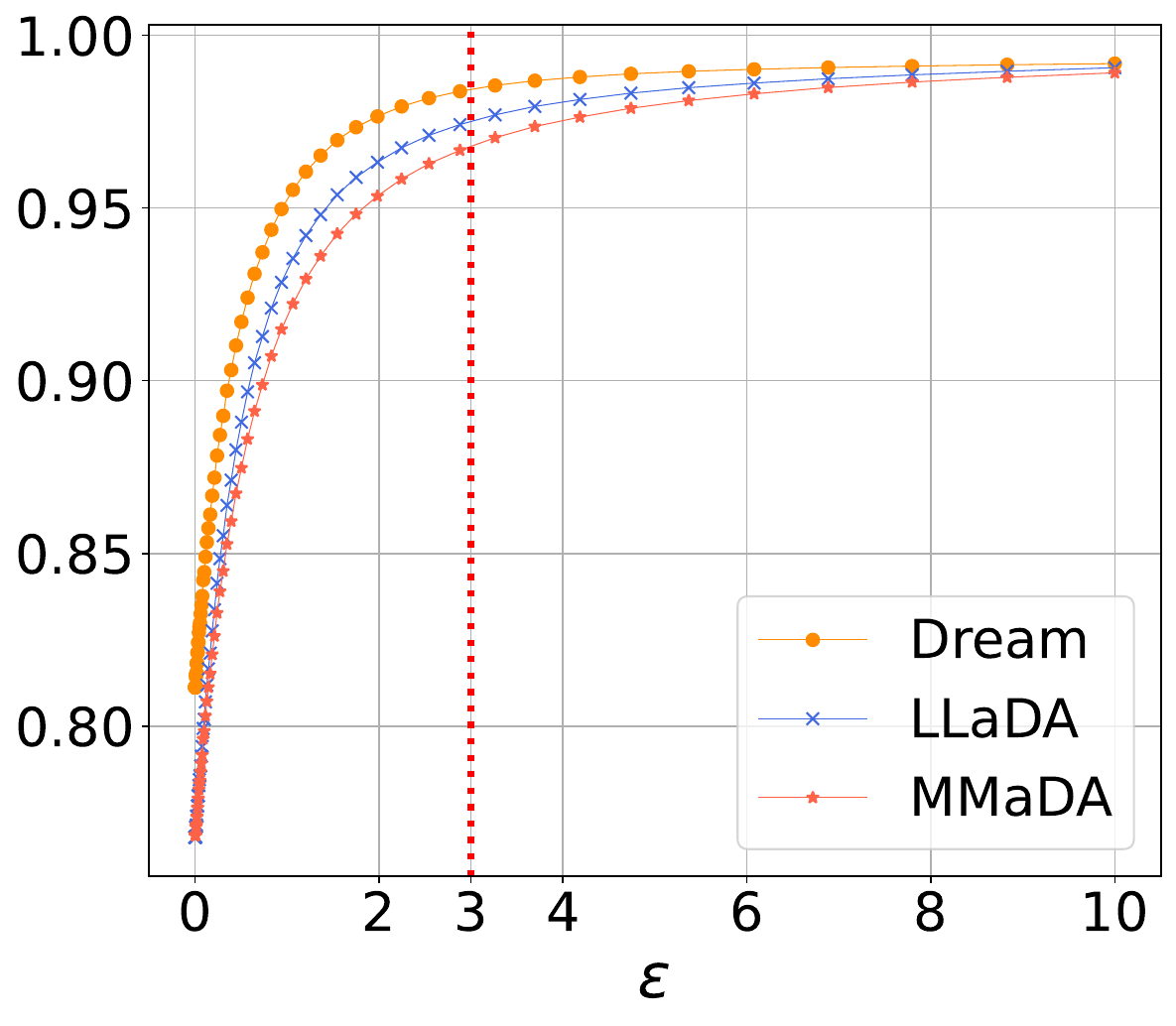}
    \hfill
    \caption{Percentage of tokens not considered as sinks when increasing the value of $\epsilon$, for a sequence of 64 tokens. A balanced threshold is found at $\epsilon = 3$, which we used in this investigation to define that a token is a sink.}
    \label{fig:avg_eps_compare}
\end{figure}

In Equation \ref{eq:metric} we defined $\epsilon$ to be the threshold for classifying a token as a sink. In \Cref{fig:avg_eps_compare} we show how the value of $\epsilon$ affects sink selection. We see that all the analysed DLMs filter out at least 96\% of tokens when using $\epsilon = 3$.   
\end{document}